\documentclass[lettersize,journal]{IEEEtran}
\usepackage{amsmath,amsfonts}
\usepackage{algorithmic}
\usepackage{array}
\usepackage[caption=false,font=normalsize,labelfont=sf,textfont=sf]{subfig}
\usepackage{textcomp}
\usepackage{stfloats}
\usepackage{url}
\usepackage{verbatim}
\usepackage{graphicx}
\usepackage{booktabs}
\usepackage{color}
\usepackage{multirow}
\makeatletter
\let\NAT@parse\undefined
\makeatother
\usepackage{hyperref}  %hyperref still needs to be put at the end!
\def\BibTeX{{\rm B\kern-.05em{\sc i\kern-.025em b}\kern-.08em
    T\kern-.1667em\lower.7ex\hbox{E}\kern-.125emX}}
\usepackage{balance}

\begin{document}

\title{Augmentation-Free Graph Contrastive Learning of Invariant-Discriminative Representations}

\author{Haifeng Li, Jun Cao, Jiawei Zhu*, Qinyao Luo, Silu He, Xuyin Wang% <-this % stops a space
\thanks{}% <-this % stops a space

\thanks{This paper is supported by National Natural Science Foundation of China (42271481, 41871364) and Hunan Provincial Natural Science Foundation of China (2022JJ30698).}

\thanks{H. Li, J. Cao, J. Zhu, Q. Luo, S. H, and X. Wang are with School of Geosciences and Info-Physics, Central South University (Corresponding author J. Zhu, Email: jw\_zhu@csu.edu.cn).}% <-this % stops a space
\thanks{}
}

\maketitle

\begin{abstract}
Graph contrastive learning is a promising direction toward alleviating the label dependence, poor generalization and weak robustness of graph neural networks, learning representations with invariance, and discriminability by solving pretasks. The pretasks are mainly built on mutual information estimation, which requires data augmentation to construct positive samples with similar semantics to learn invariant signals and negative samples with dissimilar semantics in order to empower representation discriminability. However, an appropriate data augmentation configuration depends heavily on lots of empirical trials such as choosing the compositions of data augmentation techniques and the corresponding hyperparameter settings. We propose an augmentation-free graph contrastive learning method, invariant-discriminative graph contrastive learning (iGCL), that does not intrinsically require negative samples. iGCL designs the invariant-discriminative loss (ID loss) to learn invariant and discriminative representations. On the one hand, ID loss learns invariant signals by directly minimizing the mean square error between the target samples and positive samples in the representation space. On the other hand, ID loss ensures that the representations are discriminative by an orthonormal constraint forcing the different dimensions of representations to be independent of each other. This prevents representations from collapsing to a point or subspace. Our theoretical analysis explains the effectiveness of ID loss from the perspectives of the redundancy reduction criterion, canonical correlation analysis, and information bottleneck principle. The experimental results demonstrate that iGCL outperforms all baselines on 5 node classification benchmark datasets. iGCL also shows superior performance for different label ratios and is capable of resisting graph attacks, which indicates that iGCL has excellent generalization and robustness. The source code is available at https://github.com/lehaifeng/T-GCN/tree/master/iGCL.
\end{abstract}


% \fi
\IEEEpeerreviewmaketitle

\section{Introduction}

\IEEEPARstart{G}{raph} contrastive learning (GCL) is a promising method that addresses the label dependence, poor generalization, and weak robustness of graph neural networks (GNNs). In general, contrastive learning is mainly built on the idea of mutual information estimation, which learns invariant and discriminative representations. Contrastive learning can be interpreted as a powerful discriminator that determines whether a pair of samples is derived from a joint distribution (positive samples) or from two marginal distributions (negative samples) \cite{hjelm2018learning}. Contrastive learning pulls the positive samples with similar semantic information close to capture the invariant signal, and negative samples with dissimilar semantic information are pushed away to guarantee the discriminability of representations. Naturally, the quality of the positive/negative samples is crucial for the performance of GCL \cite{you2020graph}.  

GCL relies heavily on empirical data augmentation to construct positive and negative samples. In computer vision, positive samples are generated by randomly cropping, resizing, and color distorting the target image, and negative samples are randomly sampled from the rest of the images \cite{chen2020simple}. Similar to the data augmentation of images, the common data augmentation techniques of graphs are node feature masking, node feature shuffle, edge modification, graph diffusion, and subgraph sampling \cite{liu2022graph}. GRACE \cite{zhu2020deep} and CCA-SSG \cite{zhang2021canonical} generate positive samples by randomly masking node features and removing edges. MVGRL \cite{hassani2020contrastive} proposes two graph diffusion methods based on Personalized PageRank \cite{page1999pagerank} and heat kernel \cite{kondor2002diffusion} to capture the global structural information and treats the diffuse graphs as positive samples. Benefiting from the adjacency of graphs, SUBG-CON \cite{jiao2020sub} proposes a subgraph contrastive learning method that uses subgraphs of the target node as positive samples. Meanwhile, negative samples are generated in various ways including row-wise perturbing node features \cite{velickovic2019deep} and random sampling from the remaining nodes \cite{zhu2020deep}. You et al. \cite{you2020graph} systematically analyzed the impact of four data augmentation techniques and their compositions on downstream tasks and summarized some observations. However, determining the type of data augmentation suitable for a particular task still relies heavily on empirical trials. Meanwhile the hyperparameters of data augmentation exponentially expand the search space of data augmentation configurations. Therefore, an applicable data augmentation configuration consumes a significant amount of time and computation resources.  

We rethink the essence of GCL to address the problem of empirical data augmentation. For GCL, positive samples are used to capture the invariant signal of data, and negative samples are used to prevent the representation from being indistinguishable \cite{wang2020understanding}. We separately analyze the role of positive and negative samples in order to discard data augmentation. For positive samples, data augmentation can be considered a transformation function that does not destroy the signal inherent in the data. Inspired by Siamese networks \cite{bromley1993signature}, we argue that two neural networks with similar or the same weights can transform the target sample into two representations with similar semantic information. Kefato et al. \cite{kefato2021jointly} used two projectors to transform the embeddings extracted by GNNs in order to obtain two representations with similar semantic information. In addition, neighboring nodes may have similar semantic information as the target node based on the homophily assumption  \cite{mcpherson2001birds, grover2016node2vec}. For example, neighboring nodes are often classified in the same community as the target node in community detection. We argue that neighboring nodes can potentially serve as positive samples. Many loss functions, such as the mean square error (MSE), can pull the positive samples close. The role of negative samples is to avoid the representations from collapsing into a single point or a subspace, ensuring that the representation is discriminative \cite{wang2020understanding}. To achieve this goal, a considerable number of negative samples must be generated \cite{he2020momentum}, which may lead to false negative samples \cite{rottmann2020detection} and exacerbate the consumption of computational resources. Recently, some studies addressed the problem of negative sampling by modifying the model architecture \cite{grill2020bootstrap}. According to the above analysis, we try to replace the empirical positive samples with Siamese networks and local structural information, and guarantee the discriminative power of representations by some effective measures.  

Here, we propose an augmentation-free GCL method called the invariant-discriminative graph contrastive learning (iGCL), which requires neither empirical data augmentation nor negative samples. Following the idea of Siamese networks, iGCL consists of two GNNs (online network and target network) with the same architecture that can generate high-quality positive samples with similar semantic information. We propose a positive sample construction strategy that selects the $K$ most similar representations from the 1-hop neighboring nodes of the target nodes as positive samples. To train the iGCL, we design the Invariant-Discriminative loss (ID loss) to include an invariance term and a discrimination term. To capture the invariant signal, we choose the simple but effective MSE as the invariance term to pull the positive samples close to the target sample. To empower representations discriminability, we add an orthonormal constraint as the discrimination term, which forces the different dimensions of the representations to be independent of each other and prevents the representation space from collapsing. In addition, our theoretical analysis explains the effectiveness of ID loss from the perspectives of redundancy reduction, canonical correlation analysis, and the information bottleneck principle.  

We conduct numerous experiments to illustrate the superior performance of the iGCL. iGCL not only outperforms supervised GCN and GAT but also all baselines on 5 node classification benchmark datasets. We also evaluate the generalization and robustness of the iGCL. The experimental results show that iGCL still achieves excellent performance at label ratios between $0.5\%$ and $20\%$ and alleviates the impact of graph attacks. Moreover, we visualize the representations by t-SNE, which intuitively illustrates the discriminative power. Finally, the experimental results of hyperparameter sensitivity analysis illustrate the effectiveness of the positive sample construction strategy and the discrimination term. Our main contributions are as follows. 

\begin{itemize}
    \item We propose iGCL, which does not require empirical data augmentation or negative samples, to generate invariant and discriminative representations.
    \item We show that ID loss is equivalent to the redundancy reduction criterion and canonical correlation analysis under specific conditions, and iGCL can be considered an instance of the information bottleneck principle under self-supervised learning.
    \item Experimental results show that iGCL outperforms all baselines on 5 node classification benchmark datasets and has superior generalization and robustness.
\end{itemize} 

The remainder of the paper is organized as follows: in Section 2, we briefly review related work on GCL. Section 3 describes the network architecture, positive sample construction strategy, and ID loss in detail. Section 4 evaluates the performance, generalization, and robustness of iGCL through extensive experiments. Section 5 concludes this paper.

\section{Related works}

\textbf{Graph contrastive learning.} Inspired by the great success of contrastive learning in computer vision, many works have recently made significant progress in adapting it to graphs. With regard to structure, graph contrastive learning (GCL) develops cross-scale and same-scale contrastive learning. The typical cross-scale GCL is DGI \cite{velickovic2019deep}. Following the idea of Deep InfoMax \cite{hjelm2018learning}, DGI attempts to train itself by maximizing the mutual information between local patch (node-level) representations and high-level global (graph-level) representations, capturing globally relevant information and avoiding overemphasis on proximity. HDGI \cite{ren2019heterogeneous} and ConCH \cite{li2021leveraging} further generalize DGI to heterogeneous graphs by aggregating node representations based on different types of edges. MVGRL \cite{hassani2020contrastive} also takes cross-scale contrastive learning and introduces two graph diffusion approaches to generate similar semantic samples. In addition, some works use same-scale contrastive learning. Inspired by SimCLR \cite{chen2020simple}, GRACE \cite{zhu2020deep} constructs two augmented views by randomly removing edge and mask node features. Then, GRACE learns the informative representations by reducing the distance between node representations of two views and pushing away other node representations. Since uniformly removing edges and shuffling features may lead to suboptimal augmentation, GCA \cite{zhu2021graph} improved GRACE by using an adaptable augmentation technique. To address the problem of negative sample sampling, BGRL \cite{thakoor2021bootstrapped} avoids negative samples by adopting a network architecture similar to BYOL \cite{grill2020bootstrap}. CCA-SSG \cite{zhang2021canonical} proposes a loss function based on canonical correlation analysis, which ensures that it no longer requires the parameterized mutual information estimator, additional projector, asymmetric structures, and negative samples. AFGRL \cite{lee2022augmentation} devolops an augmentation-free GCL method. AFGRL utilizes the nodes that share local structural information and global semantic information for target nodes as positive samples. iGCL differs from AFGRL in three ways: 1. the ID loss differs from the loss of AFGRL, 2. iGCL generates positive samples in a more efficient way than AFGRL without losing performance, and 3. iGCL can be applied to large-scale graphs with the help of neighbor sampling, while AFGRL needs to load entire graphs into memory.  

\textbf{Data augmentation on graphs.} In computer vision, SimCLR \cite{chen2020simple} states that the composition of data augmentations plays a critical role in contrastive learning. GraphCL \cite{you2020graph} empirically illustrates that data augmentation is also crucial for GCL and that the compositions of different augmentation methods are beneficial for improving performance through extensive experiments. Liu et al. \cite{liu2022graph} systematically categorized the data augmentation on graphs into node feature masking, node feature shuffle, edge modification, graph diffusion, and subgraph sampling. For example, GRACE \cite{zhu2020deep} and CCA-SSG \cite{zhang2021canonical} generate positive samples by randomly masking node features and removing edges. However, GraphCL indicates that different types of graphs are suitable for different kinds of data augmentation. The choice of data augmentation for a given task relies heavily on trial and error, which is certainly time-consuming and computationally resource-intensive. Meanwhile, the hyperparameters of data augmentations exponentially expand the search space. We try to generate positive samples based on the intrinsic signal of graphs instead of empirical data augmentation.

\section{Methodology}
In this section, we propose a simple but effective GCL method called invariant-discriminative graph contrastive learning (iGCL). iGCL utilizes the Siamese network architecture to generate positive samples in the representation space. In addition, the positive sample construction strategy leverages the similarity of representations and local structural information to enhance the quality and diversity of positive samples. ID loss not only ensures learning invariant representations but also avoids trivial solutions.

\subsection{iGCL framework}

\begin{figure}[h]
  \begin{center}
  \includegraphics[width=3.3in]{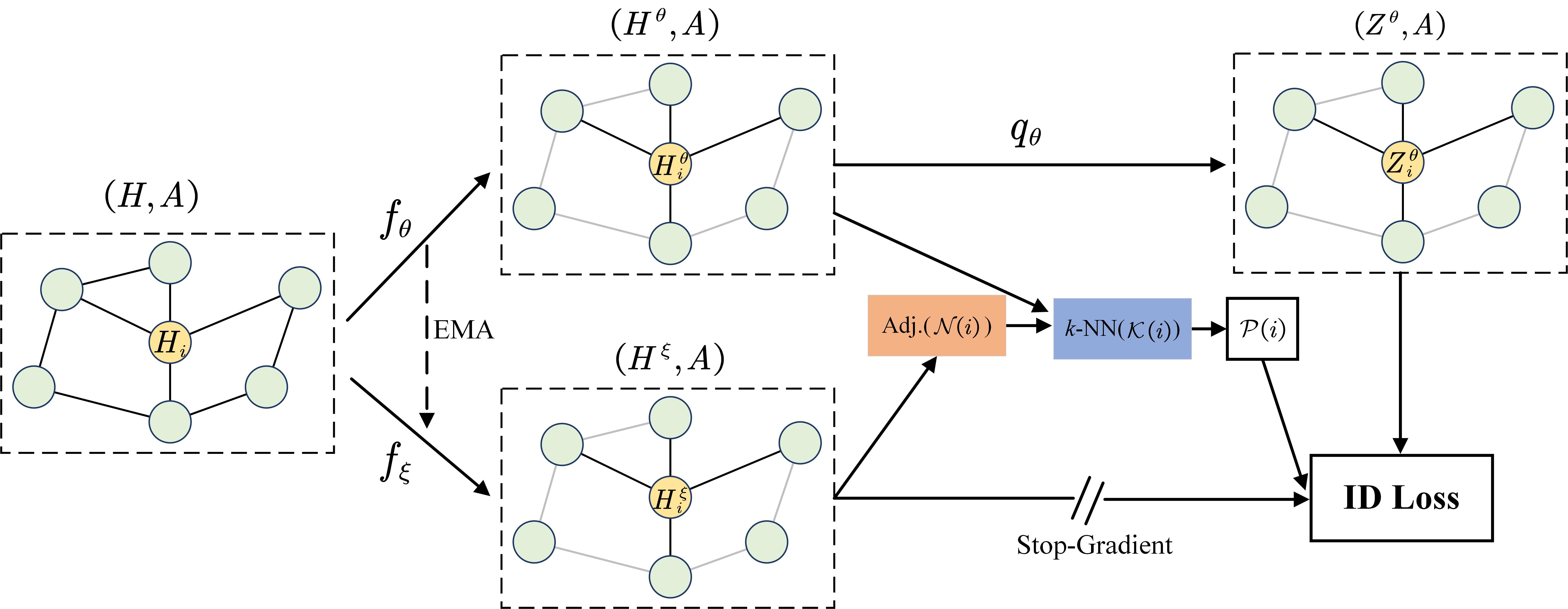}
  \end{center}
  \caption{Network architecture of iGCL.}
  \label{fig:network architecture}
\end{figure}

The network architecture of iGCL is shown in Fig. \ref{fig:network architecture}. iGCL consists of an online network $f_{\theta}$, a target network $f_{\xi}$ and a projector $q_{\theta}$ . The learnable parameters $\theta, \xi$ of $f_{\theta}$ and $f_{\xi}$ are shared, which can transform a sample into two representations with similar semantic information. Suppose we have a graph $G=(V,E)$ with features on the nodes $H=(h_1, h_2,...,h_N), h_i \in \mathbb{R}^F$. $N=|V|$ is the number of nodes, and $F$ is the dimension of features. iGCL generates the online representations and the target representations $H^{\theta}, H^{\xi} \in \mathbb{R}^{N \times D}$, $D$ is the dimension of representations, by inputting $H$ and the adjacency matrix $A$ into $f_{\theta}$ and $f_{\xi}$, respectively. Then, iGCL transforms $H^{\theta}$ into $Z^{\theta} \in \mathbb{R}^{N \times D^q}$ by $q_{\theta}$, $D_q$ is the dimension of $z_i^{\theta}$. Next, the positive sample construction strategy automatically finds the qualified positive sample representations from $H^{\xi}$. Finally, ID loss guarantees the ability to learn the invariant and discriminative $H^{\theta}$. Intriguingly, the experimental results of Hyperparameter sensitivity analysis suggest that $H^{\theta}$ benefits from extremely high dimension.

Now, we elaborate on how the learnable parameters $\theta, \xi$ are shared. When initializing $f_{\theta}, f_{\xi}$ , the learnable parameters are identical, i.e., $\theta = \xi$. During training, $\theta$ is updated according to ID loss, while $\xi$ is updated by the exponential moving average (EMA) technique. EMA can be formulated as Eq. \ref{eq: ema}, where $\tau \in [0, 1]$ is the balance coefficient. EMA can improve the effectiveness of the Siamese network in contrastive learning \cite{grill2020bootstrap}.

\begin{equation}
    \label{eq: ema}
    \xi \gets \tau \xi + (1 - \tau) \theta
\end{equation}

\subsection{Learning invariant signal}

\subsubsection{Positive sample construction strategy}
\begin{figure}[h]
  \begin{center}
  \includegraphics[width=3.5in]{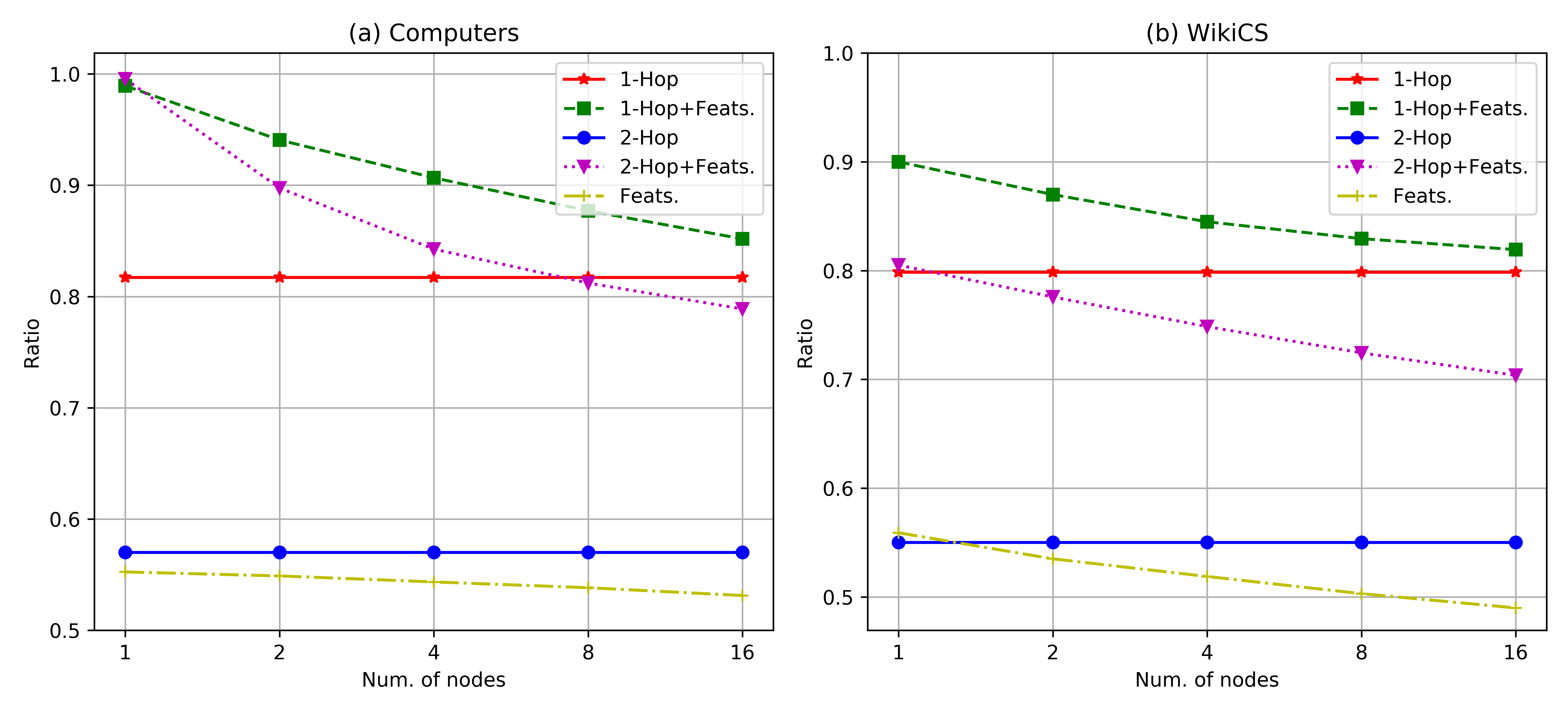}
  \end{center}
  \caption{Proportion of nodes with same class of target nodes obtained by different construction strategies. Num of nodes indicates number of nodes with most similar features to target node.}
  \label{fig:neighbors_new}
\end{figure}

iGCL utilizes the similarity of representations and the local structure to construct multiple positive samples of target nodes. Benefiting from the Siamese network, the representation $H_i^{\xi}$ can be considered as the direct positive sample set $\{i\}$ of the node $i$. Besides, the positive sample construction strategy enriches the diversity of positive samples. We argue that the neighboring nodes connected to the target nodes potentially share similar semantic information. For example, the similar semantic information of nodes can be interpreted as having the same labels for the node classification task. The red line (1-hop neighboring nodes) in Fig. \ref{fig:neighbors_new} indicates that the neighboring nodes tend to have the same label as the target node, with the proportion exceeding even 80$\%$. Then, we explore the correlation between feature similarity and semantic similarity. We choose the cosine similarity to measure the feature similarity. Although the yellow line (Feats.) in Fig. \ref{fig:neighbors_new} indicates that feature similarity is not highly correlated with semantic similarity, feature similarity can enhance the correlation between 1-hop structural information and semantic similarity, as shown by the green line (1-Hop+Feats). As the local structure expands to a 2-hop neighbor, the proportion of neighboring nodes with the same label as the target node decreases significantly, as shown by the blue line (2-Hop) in Fig. \ref{fig:neighbors_new}. In contrast, adding the feature similarity can substantially exclude neighboring nodes with different labels from the target nodes, as shown by the purple line (2-Hop+Feats) in Fig. \ref{fig:neighbors_new}. We also observe that the proportion of nodes with the same label decreases when the number of the most similar nodes increases.  

In light of the above observations, we regard the $k$-hop neighboring nodes as the alternative positive sample set $\mathcal{N}(i)$ for target node $i$. Meanwhile, there is still a portion of $\mathcal{N}(i)$ whose semantics are not similar to those of $i$. We select the $k$-nn algorithm to filter out these nodes and regard the $k$ most similar representations as the supplementary positive sample set $\mathcal{K}(i)$. Given that ID loss is based on the MSE, the shorter the Euclidean distance of the representations, the more similar their semantic information is likely to be. Therefore, the Euclidean distance is chosen as the metric function of $k$-nn. Finally, the total positive sample set of target node $i$ is $\mathcal{P}(i)=(\{i\} \cup \mathcal{K}(i))$. In this paper, we set the 1-hop neighboring nodes as $\mathcal{N}(i)$.

To evaluate ID loss, we need to organize $\mathcal{P}(i)$ into $K=1+k$ parts. First, we treat the direct positive sample $\{i\}$ as a part $\mathcal{P}_1(i)=i$ in which the index of positive nodes is the same as the index of target nodes. Next, we process the supplementary positive sample set $\mathcal{K}$. For node $i$, we descendingly sort the nodes in $\mathcal{K}(i)$ according to similarity. $\mathcal{K}$ is divided into $\mathcal{P}_2(i), \mathcal{P}_3(i), \mathcal{P}_K(i)$, for a total of $K-1$ parts. Given $1<n<m\leq K$, the similarity between node $\mathcal{P}_n(i)$ and node $i$ is always greater than that of node $\mathcal{P}_m(i)$. Note that the number of positive samples of node $i$ may be less than $K$ if the degree of node $i$ is less than $K-1$. For example, node $i$ does not exist in $\mathcal{P}_4$ when $K=4$ and $D_{ii}=2$. In other words, node $i$ is not involved in calculating $\mathcal{P}_4$ of the ID loss.

\subsubsection{Invariant Loss}
To maximize the invariant signal, we simply but effectively minimize the MSE between the target samples and positive samples in the representation space. The representations are standardized as $\bar{Z}=(Z-\mu (Z))/(N \times \sigma(Z))$ to eliminate the effect of scale. For standardized $\bar{Z}^{\theta}, \bar{H}^{\xi}$, the invariant loss is formulated as Eq. \ref{eq: Invariant Loss}.

\begin{equation}
    \label{eq: Invariant Loss}
    \mathcal{L}_{\theta, \xi}=\left\|\bar{Z}^\theta-\bar{H}^{\xi}\right\|_F^2
\end{equation}

Unfortunately, Eq. (2) allows the iGCL to obtain a trivial solution, i.e., the representations collapse into a subspace or point. For example, $\bar{Z}^{\theta}, \bar{H}^{\xi}$ are the same constant.

\subsection{Guaranteeing discriminability of representations}

Inspired by Laplacian Eigenmaps \cite{belkin2003laplacian}, we guarantee the discriminability of representations by an orthonormal constraint. The orthonormal constraint ensures that the different dimensions of $\bar{Z}^{\theta}, \bar{H}^{\xi}$ are linearly independent of each other. In other words, the orthonormal constraint ensures no redundant dimensions for $\bar{Z}^{\theta}, \bar{H}^{\xi}$. Furthermore, $\bar{H}^{\theta}$ avoids collapsing into a subspace or point. To satisfy the above requirement, Eq. \ref{eq: Invariant Loss} is modified to the invariant loss with an orthonormal constraint, as shown in Eq. \ref{eq: Guaranteeing discriminability of representations}.

\begin{equation}
    \label{eq: Guaranteeing discriminability of representations}
    \begin{split}
        & \mathcal{L}_{\theta, \xi}=\left\|\bar{Z}^\theta-\bar{H}^{\xi}\right\|_F^2 \\
        & s.t. \ \bar{Z}^{\theta \mathrm{T}} \bar{Z}^\theta=I \  and \  \bar{H}^{\xi \mathrm{T}} \bar{H}^{\xi}=I \\
    \end{split}
\end{equation}

\begin{figure}
    \begin{center}
        \includegraphics[width=3.5in]{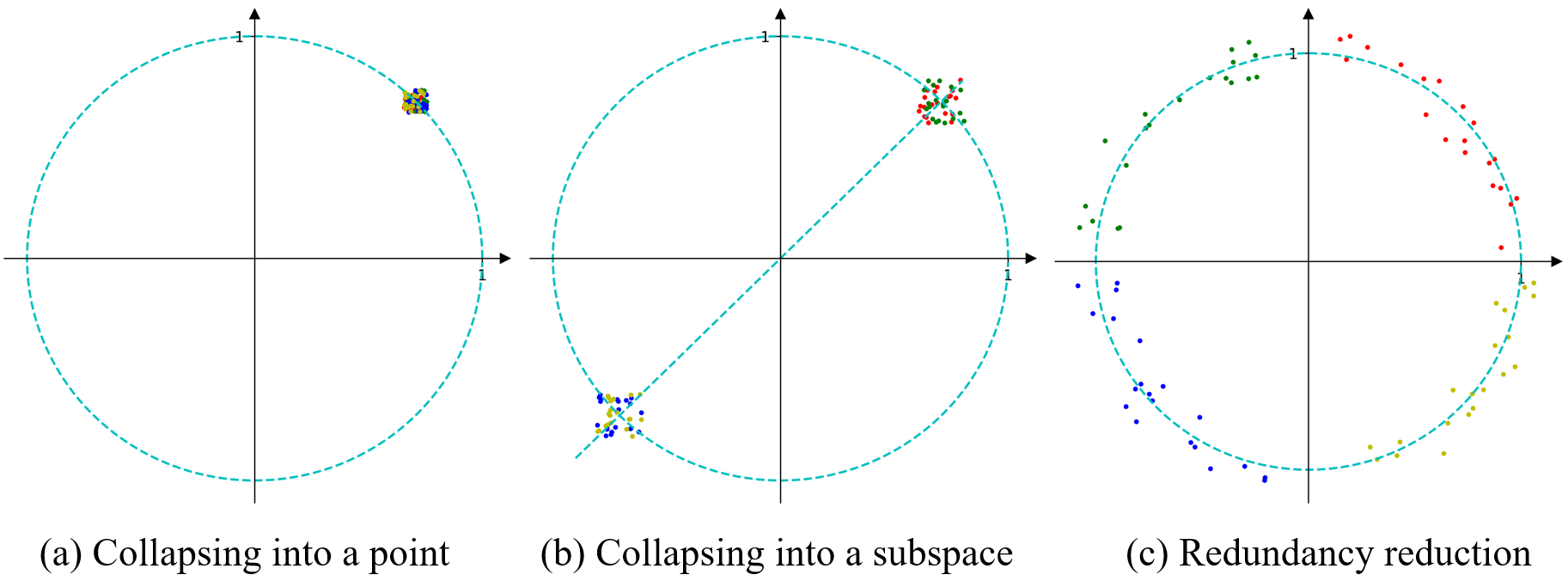}
    \end{center}
    \caption{Illustration of effect of orthonormal terms. Coordinates of nodes represent two-dimensional representations. Nodes of same color are semantically similar.}
    \label{fig: illu of loss function}
\end{figure}

Eq. \ref{eq: Guaranteeing discriminability of representations} encourages $\bar{Z}^{\theta}, \bar{H}^{\xi}$ to be as similar as possible, while the orthonormal constraint ensures that different dimensions of representations are linearly independent of each other. We illustrate the utility of the orthonormal constraint through \ref{fig: illu of loss function}. If the representations collapse to a single point, we are completely unable to distinguish semantically dissimilar nodes, as shown in Fig. \ref{fig: illu of loss function}-a. If the representations collapse into a subspace, we have difficulty distinguishing nodes with similar representations but dissimilar semantic information, such as those points at two endpoints of the line in Fig. \ref{fig: illu of loss function}-b. It wastes a dimension and impairs the discriminability of representations. In contrast, the orthonormal constraint ensures that iGCL takes advantage of the dimensionality of the representations, equipping the representations to discriminate semantically dissimilar nodes, as shown in Fig. \ref{fig: illu of loss function}-c.

In addition, the architecture of iGCL has the ability to resist a trivial solution. iGCL employs tricks such as asymmetric structures and an additional projector. Both Eq. \ref{eq: Guaranteeing discriminability of representations} and the architecture guarantee that the representations have the ability to discriminate semantically dissimilar nodes.

\subsection{Invariant-discriminative loss}

Based on the above analysis, we introduce the loss function of iGCL, invariant-discriminative loss (ID loss). To reduce the computational complexity, we use the Lagrange multiplier to relax the invariant loss with the orthonormal constraint into an unconstrained objective function, as shown in Eq. (4), where $\lambda$ is the balance coefficient. We refer to Eq. \ref{eq: Invariant-discriminative loss} as ID loss.

\begin{equation}
    \label{eq: Invariant-discriminative loss}
    \begin{split}
        \mathcal{L}_{\theta, \xi} = & \underbrace{\left\|\bar{Z}^\theta-\bar{H}^{\xi}\right\|_F^2}_{\text {invariance term}}+ \\
         & \underbrace{\lambda ( \left\|\bar{Z}^{\theta \mathrm{T}} \bar{Z}^\theta-I\right\|_F^2+\left\|\bar{H}^{\xi \mathrm{T}} \bar{H}^{\xi}-I\right\|_F^2 )}_{\text {discrimination term }}
    \end{split}
\end{equation}

ID loss consists of an invariance term and a discrimination term. The invariance term forces iGCL to learn the invariant signal of datasets. The discrimination term forces different dimensions of the representations to be linearly independent of each other, which guarantees the discriminability of representations.

We further modify the ID loss to fit the positive sample construction strategy. The final loss is the mean ID loss of the $K$ parts of $\mathcal{P}$, formulating as Eq. \ref{eq: Invariant-discriminative loss with neighbors}, where $\bar{Z}^{\theta} [ \mathcal{P} ]$ indicates $\bar{Z}^{\theta}$ arranged by the index of $\mathcal{P}_k$. For $j \notin \mathcal{P}_k(i)$ , we skip the representations of the corresponding rows in $\bar{Z}^{\theta}, \bar{H}^{\xi}$. For conciseness, we do not mark the case in Eq. \ref{eq: Invariant-discriminative loss with neighbors}.

\begin{equation}
    \label{eq: Invariant-discriminative loss with neighbors}
    \begin{split}
            \mathcal{L}_{\theta, \xi} = & \frac{1}{K} \sum_{k=1}^K \left\|\bar{Z}^\theta\left[\mathcal{P}_k\right]-\bar{H}^{\xi}\right\|_F^2+ \\
            & \lambda ( \left\|\bar{Z}^\theta\left[\mathcal{P}_k\right]^{\mathrm{T}} \bar{Z}^\theta\left[\mathcal{P}_k\right]-I\right\|_F^2+\left\|\bar{H}^{\xi \mathrm{T}} \bar{H}^{\xi}-I\right\|_F^2 ) 
    \end{split}
\end{equation}

\subsection{Theoretical analysis of ID loss}

Different from instance-level mutual information estimation, ID loss is feature-level, which is particularly evident at the discrimination term. Mutual information estimation is generally achieved by pushing away negative samples, while ID loss is achieved by forcing the different dimensions of the representations to be linearly independent of each other. Naturally, iGCL does not require negative samples. We further show that ID loss is theoretically connected with the redundancy reduction criterion, canonical correlation analysis, and information bottleneck principle. For the sake of analysis, we do not consider the multiple positive samples generated by the positive sample sampling strategy.

\subsubsection{Connection with redundancy reduction criterion}
Neuroscientist H. Barlow proposed the redundancy reduction criterion \cite{barlow1961possible}, which argues that the goal of sensory processing is to recode highly redundant sensory input into a factorial code (each of its parts is statistically independent). Zbontar et al. \cite{zbontar2021barlow} applied the redundancy reduction criterion to self-supervised learning in computer vision, which trains models by a loss function based on the cross-correlation matrix between the representations of two views. The loss function is shown in Eq. \ref{eq: redundancy reduction criterion}, where $\mathcal{C} = \bar{Z}^{\theta T} \bar{H}^{\xi}$ is a covariance matrix.

\begin{equation}
    \label{eq: redundancy reduction criterion}
    \mathcal{L}=\underbrace{\sum_i\left(1-\mathcal{C}_{i i}\right)^2}_{\text {invariance term }}+\underbrace{\lambda \sum_i \sum_{j \neq i} \mathcal{C}_{i j}^2}_{\text {redundancy reduction term }}
\end{equation}

The purpose of this loss function is to make representations in the same dimension of two views linearly correlated in order to learn invariant signals and to let representations in different dimensions be linearly uncorrelated to avoid representations collapsing to a subspace or point. Following the redundancy reduction criterion, ID loss tries to linearly correlate the same dimension and linearly uncorrelated different dimensions of $\bar{Z}^{\theta}, \bar{H}^{\xi}$, respectively. ID loss uses the MSE instead of the invariance term of Eq. \ref{eq: Invariant-discriminative loss}.

\subsubsection{Connection with canonical correlation analysis}

Canonical correlation analysis (CCA) is applied to measure the correlation of multivariate variables. The goal is to find two vectors $a \in \mathbb{R}^m, b \in \mathbb{R}^n$ to reduce the dimensionality of two multivariate variables $X \in \mathbb{R}^m, Y \in \mathbb{R}^n$ into a univariate variable, and maximize the correlation coefficient $\rho=\operatorname{corr}\left(a^{\mathrm{T}} X, b^{\mathrm{T}} Y\right)=\frac{a^{\mathrm{T}} \Sigma_{X Y} b}{\sqrt{a^{\mathrm{T}} \Sigma_{X X} a} \sqrt{b^{\mathrm{T}} \Sigma_{Y Y} b}}$, where $\sum_{X Y}=Cov(X,Y)$ is a covariance matrix. CCA is equivalent to Eq. \ref{eq: CCA}.

\begin{equation}
    \label{eq: CCA}
    \max _{a, b} \quad a^{\mathrm{T}} \Sigma_{X Y} b, \text { s.t. } a^{\mathrm{T}} \Sigma_{X X} a=1, b^{\mathrm{T}} \Sigma_{Y Y} b=1
\end{equation}

Eq. \ref{eq: CCA} clearly illustrates that CCA attempts to maximize the correlation of two multivariate variables with the linear transformation while making the different variables independent of each other. \cite{andrew2013deep, gong2014multi} introduced CCA to multi-view deep learning by replacing the linear transformation with neural networks. Soft CCA \cite{chang2018scalable} further removes the hard decorrelation constraint by Lagrange multiplier, which saves a lot of computational resources. The final soft CCA is Eq. \ref{eq: soft CCA}, where $P_{\theta_1}, P_{\theta_2}$ are two neural networks, $X_1, X_2$ are two views of a sample, $\mathcal{L}_{dist}$ measure the correlation of $P_{\theta_1}(X_1), P_{\theta_2}(X_2)$, $\mathcal{L}_{SDL}$ (stochastic decorrelation loss) computes the distance between $P_{\theta_i}(X_i)^T P_{\theta_i}(X_i)$ and the identity matrix $I$, for $i=1,2$.

\begin{equation}
    \label{eq: soft CCA}
    \begin{split}
         \min _{\theta_1, \theta_2} \quad
         & \mathcal{L}_{\text {dist }}\left(P_{\theta_1}\left(X_1\right), P_{\theta_2}\left(X_2\right)\right)+ \\
         & \lambda(\mathcal{L}_{S D L}\left(P_{\theta_1}\left(X_1\right)\right)+\mathcal{L}_{S D L}\left(P_{\theta_2}\left(X_2\right)\right))
    \end{split}
\end{equation}

$\bar{Z}^{\theta}, \bar{H}^{\xi} \in \mathbb{R}^{N \times F}$ can be viewed as $N$ samples by randomly sampling from two multivariate variables $X, Y \in \mathbb{R}^F$, respectively. When $\mathcal{L}_{dist}, \mathcal{L}_{SDL}$ is the MSE, Eq. \ref{eq: soft CCA} is the loss function of CCA-SSG, which is equivalent to ID loss. The insight of CCA-SSG is derived from CCA, while the insight of iGCL is derived from the analysis of the invariance and discriminability of the representations.

\subsubsection{Connection with information bottleneck principle}

iGCL is an instance of the Information Bottleneck (IB) principle under self-supervised learning setting. We further clarify the connection between ID loss and IB. The self-supervised IB is developed from the supervised IB. The supervised IB can be formulated as Eq. \ref{eq: IB}, where $X$ indicates the random variable of input data, $Y$ indicates the labels of downstream tasks, $Z_X$ indicates the representations of $X$, $I(\dot)$ indicates the mutual information. 

\begin{equation}
    \label{eq: IB}
    \mathcal{I B}_{\text {sup }}=I\left(Y, Z_X\right)-\beta I\left(X, Z_X\right), \text { where } \beta>0
\end{equation}

$\mathcal{I B}$ tries to maximize the mutual information between $Z_X$ and $Y$, while minimize the mutual information between $Z_X$ and $Y$. The aim of $\mathcal{I B}$ is that $Z_X$ only preserve the useful information for predicting $Y$. Some works \cite{amjad2019learning, federici2020learning} develop supervised IB to self-supervised IB. self-supervised IB is formulated as Eq. \ref{eq: self_supervised IB}, where $S$ indicates self-supervised signals (the augmented views of $X$), $Z_S$ indicates the representation of $S$.

\begin{equation}
    \label{eq: self_supervised IB}
    \mathcal{I B}_{s s l}=I\left(X, Z_S\right)-\beta I\left(S, Z_S\right), \text { where } \beta>0
\end{equation}

Naturally, $\mathcal{I B}_{s s l}$ tries to maximize to the mutual information between $Z_S$ and $S$ and to expect $Z_S$ is the maximally compressed representation of $X$. Zhang et al. \cite{zhang2021canonical} prove that minimizing ID loss is equivalent to maximize $\mathcal{I B}_{s s l}$ under $0<\beta<1$, i.e., $\min \mathcal{L}_{ID \  Loss} \Rightarrow \max \mathcal{I B}_{s s l}$. It suggests that iGCL can be viewed as an instance of IB under self-supervised learning setting. Assuming that data augmentation does not change label-related information, it means that all the task-relevant information is preserved in augmentation invariant features. Minimizing ID loss can maximally preserving the task-relevant information while reducing the task-irrelevant information \cite{zhang2021canonical}.

We show that ID loss is deeply connected to the redundancy reduction criterion and CCA, which explains the significance of ID loss from different perspectives. With the help of \cite{zhang2021canonical}, we are aware that minimizing ID loss is equivalent to maximizing the self-supervision IB under a specific condition. Moreover, ID loss ensures that the learned representations are expected to retain minimal but sufficient information about downstream tasks, which explains why the representations generated by iGCL can perform well on downstream tasks.

\section{Experiments and results}

\subsection{Datasets}

\begin{table}[h]
    \centering
    \caption{Statistical details of node classification benchmark datasets}
    \label{tab: details of datasets}
    \setlength{\tabcolsep}{1.1mm}{
        \begin{tabular}{lllllll}
            \hline
             & Nodes & Edges & Features & Classes & Directed &  Avg Degree \\
            \hline 
            CS & 18333 & 10027 & 6805 & 15 & False & 8.93 \\
            Physics & 34493 & 282455 & 8415 & 5 & False & 14.38 \\
            Computers & 13381 & 259159 & 757 & 10 & False & 35.76 \\
            Photo & 7487 & 126530 & 745 & 8 & False & 31.13 \\
            WikiCS & 11701 & 216123 & 300 & 10 & True & 36.94 \\
            \hline
        \end{tabular}
    }
\end{table}

To evaluate the performance of iGCL, we conduct extensive experiments on five node classification benchmark datasets: Coauther CS (CS), Coauthor Physics (Physics), Amazon Computers (Computers), Amazon Photo (Photo), and WikiCS. CS, Physics, Computers, and Photo are undirected graphs, while WikiCS is a directed graph. The statistical information of the datasets is listed in Table \ref{tab: details of datasets}. Details of the datasets are as follows.

\begin{itemize}
    \item CS and Physics are academic networks cut from the Microsoft Academic Graph \cite{sinha2015overview}, which contains 
coauthorship relationships. In these two graphs, nodes denote authors, edges denote collaborations, labels denote authors' research fields, and node features denote a bag-of-words representation of the paper keywords.
    \item Computers and Photos are subgraphs in the Amazon copurchase relationship graph \cite{mcauley2015image}, where nodes represent items, edges represent two items that are often simultaneously purchased, and node features represent word vectors of item reviews.
    \item WikiCS \cite{mernyei2020wiki} is a web network formed by computer science-themed pages on Wikipedia with nodes representing pages, edges representing link relationships between pages, node features being word vectors extracted from paper titles and abstracts, and labels representing the disciplines of computer science.
\end{itemize}

As in \cite{lee2022augmentation}, we randomly divide the dataset into training sets, validation sets, and test sets in a ratio of 1:1:8. We repeated the random partitioning 20 times for each dataset. Since WikiCS provides 20 accepted divisions of the training and validation sets, we directly utilize its given divisions. We use the mean classification accuracy and standard deviation under the 20 different divisions as the main metric to evaluate the performance of the models.

\subsection{Baselines}

To comprehensively and fairly evaluate the performance of iGCL, we selected three categories of methods as baselines: unsupervised learning, supervised learning, and contrastive learning.  

Unsupervised learning baselines include Raw Feats and Node2Vec \cite{grover2016node2vec}. Raw Feats indicates classifying nodes using the logistic regression classifier with raw node features as input. Node2vec indicates classifying nodes using the logistic regression classifier with the representations generated by Node2vec as input.  

Supervised learning baselines include GCN \cite{kipf2016semi}, GAT \cite{velivckovic2017graph}, and GraphSAGE \cite{hamilton2017inductive}. GCN dramatically reduces the computational complexity of graph convolution through simplification operations and renormalization trick and iteratively extracts informative representations in a layer-stacked manner. GAT introduces the self-attention mechanism to GNNs to adaptively assign weights to neighboring nodes during aggregation in a data-driven manner. GraphSAGE proposes a sampling approach that allows GNNs to adapt to large-scale graphs.  

Contrastive learning baselines include DGI \cite{velickovic2019deep}, MVGRL \cite{hassani2020contrastive}, GRACE \cite{zhu2020deep}, GCA \cite{zhu2021graph}, BGRL \cite{thakoor2021bootstrapped}, CCA-SSG \cite{zhang2021canonical}, and AFGRL \cite{lee2022augmentation}. DGI and MVGRL train models by contrasting node-level representations with graph-level representations. Following the idea of SimCLR, GRACE and GCA learn node representations by pulling the representations of the same nodes in two augmented graphs closer and pushing them away from each other. Inspired by BYOL, neither BGRL nor AFGRL need negative samples to avoid a trivial solution. CCA-SSG proposes a loss function based on canonical correlation analysis to avoid a parameterized mutual information estimator, additional projector, asymmetric structures, and negative samples.

\subsection{Experimental Protocol}

We evaluate the performance of iGCL on the node classification task. As in \cite{velickovic2019deep}, we train the iGCL in an unsupervised manner. Then, we use the learned node representations and available labels to train a logistic regression classifier and use this model to predict the node classes on the test set. We use the test results when the model performs best on the validation sets.  

The encoders of the iGCL can be an arbitrary GNN. To compare fairly with the baselines, we use the GCN as the encoder of the iGCL, such as $f_{\theta}$ and $f_{\xi}$. Formally, the encoder architecture of iGCL is defined as Eq. \ref{eq: formula of GCN}, where $H^l$ indicates the node represenstations of the $l$ layer, $H^0=H$ indicates the raw node features, $\hat{A}=A+I$ indicates the adjacent matrix adding self-loop, $\hat{D}_{ii}=\sum_j \hat{A}_{ij}$ is the node degree matrix, $W^l$ indicates the learnable parameters of the $l$ layer, and $\sigma(\cdot)$ is the activate function, such as ReLU.  

\begin{equation}
    \label{eq: formula of GCN}
    \begin{split}
        H^l 
        &=GCN^l(H^{(l-1)}, A)\\
        &=\sigma(\hat{D}^{-1/2} \hat{A} \hat{D}^{-1/2} H^{(l-1)} W^l)\\  
    \end{split}
\end{equation}
	 
We use the Adam SGD optimizer with a learning rate of 0.005 and L2 regularization of 0.0001 to train the models and initialize the learnable parameters with Glorot initialization. To obtain the optimal classification accuracy, we use grid search to select the appropriate hyperparameter configuration, such as the number of layers of GCN $L$, dimension of the representations $D, D^q$, number of positive samples $K$, and coefficients of the discrimination term $\lambda$. The optimal hyperparameter configuration of the iGCL on the five benchmark datasets is shown in Table \ref{tab: hyperparameter settings}.

\begin{table}[h]
    \centering
    \caption{Hyperparameter settings of iGCL on the datasets.}
    \label{tab: hyperparameter settings}
    \setlength{\tabcolsep}{3mm}{
        \begin{tabular}{lllllll}
            \hline
             & $L$ & $D$ & $D^q$ & $K$ & $\lambda$ & Epochs\\
            \hline
            CS & 1 & 1024 & 2048 & 1 & 0.001 & 1000 \\
            Physics & 1 & 1024 & 2048 & 3 & 0.0001 & 3000 \\
            Computers & 1 & 4096 & 8192 & 4 & 0.001 & 5000 \\
            Photo & 1 & 1024 & 2048 & 4 & 0.0005 & 1000 \\
            WikiCS & 2 & 1024 & 2048 & 6 & 0.005 & 1000 \\
            \hline
        \end{tabular}
    }

\end{table}

\subsection{Comparison with Peer Methods}

We perform a fair evaluation of the performance of iGCL by comparing it with the baselines on the 5 node classification benchmark datasets. Table \ref{tab: Comparison with Peer Methods} provides statistical results in terms of the mean test set classification accuracies (in percent) and standard deviation. iGCL outperforms all types of baselines on the benchmark datasets, powerfully illustrating the superior performance of iGCL. Raw feed Node2vec always obtains the worst classification accuracies, which indicates that proper association of node features and structure is critical to improving the performance of models on graph analysis tasks. Although GNNs can utilize both node features and structure, supervised learning makes GNNs fit manual labels instead of learning the dataset's invariant representations. iGCL significantly outperforms supervised learning GNNs in terms of classification accuracies on all five benchmark datasets. In particular, iGCL improves on WikiCS by 2 percentage points. With recent advances in GCL, a range of GCLs outperform supervised learning GNNs. Compared with GCL baselines, iGCL still achieves the best classification accuracies on all of the datasets. The experimental results show that the iGCL can learn invariant and discriminative representations.

\begin{table*}
    \centering
    \caption{Summary of statistical results in terms of mean test set classification accuracies (in percent) and standard deviation on 5 node classification benchmark datasets. \textbf{Bold} numbers indicate best results.}
    \label{tab: Comparison with Peer Methods}
    \setlength{\tabcolsep}{6mm}{
    \begin{tabular}{cccccc}
        \toprule
         & CS & Physics & Computers & Photo & WikiCS\\ 
        \midrule
        GCN & 92.55Â±0.17 & 95.51Â±0.11 & 88.38Â±0.44 & 92.85Â±0.38 & 76.78Â±0.46 \\
        GAT & 92.65Â±0.51 & 94.35Â±0.40 & 88.31Â±0.64 & 92.36Â±0.59 & 76.64Â±0.51 \\
        GraphSAGE & 92.90Â±0.18 & 95.67Â±0.13 & 88.56Â±0.49 & 92.80Â±0.36 & 76.74Â±0.45 \\
        \hline
        Raw Feats. & 92.01Â±0.16 & 93.62Â±0.13 & 78.76Â±0.75 & 86.23Â±0.54 & 72.41Â±0.58 \\
        Node2Vec & 88.55Â±0.26 & 91.85Â±0.14 & 84.63Â±0.41 & 89.68Â±0.41 & 71.89Â±0.65 \\
        DGI & 92.28Â±0.16 & 94.51Â±0.52 & 87.45Â±0.46 & 91.65Â±0.32 & 74.12Â±0.40 \\
        MVGRL & 92.11Â±0.12 & 95.33Â±0.03 & 87.52Â±0.11 & 91.74Â±0.07 & 77.52Â±0.14 \\
        GRACE & 92.53Â±0.11 & 95.26Â±0.02 & 86.65Â±0.25 & 92.45Â±0.24 & 77.97Â±0.63 \\
        GCA & 92.84Â±0.14 & 95.38Â±0.05 & 87.85Â±0.31 & 92.49Â±0.33 & 77.84Â±0.67 \\
        BGRL & 92.59Â±0.14 & 95.48Â±0.08 & 89.69Â±0.37 & 92.82Â±0.38 & 76.86Â±0.74 \\
        CCA-SSG & 93.01Â±0.20 & 95.42Â±0.09 & 88.76Â±0.36 & 92.89Â±0.28 & 75.67Â±0.73 \\
        AFGRL & 93.21Â±0.14 & 95.69Â±0.11 & 89.82Â±0.40 & 92.93Â±0.26 & 77.57Â±0.45 \\
        iGCL & \textbf{93.35Â±0.14} & \textbf{95.85Â±0.10} & \textbf{90.06Â±0.41} & \textbf{93.10Â±0.26} & \textbf{78.22Â±0.69} \\
        \bottomrule
    \end{tabular}
    }
\end{table*}

\subsection{Effects of label magnitude}

We further evaluate the generalizability of the iGCL under different label ratios. In this subsection, we divide the datasets into training sets and test sets. The training set is divided into proportions {0.5$\%$, 1$\%$, 2$\%$, 5$\%$, 10$\%$, 20$\%$}, and the corresponding proportions for the test set are {99.5$\%$, 99$\%$, 98$\%$, 95$\%$, 90$\%$, 80$\%$}. Each proportion was randomly divided 20 times. Label proportions of 0.5$\%$ and 1$\%$ represent cases where labels are difficult to produce in the real world, while label proportions of 20$\%$ represent cases where labels are easy to obtain. The classification accuracies of the iGCL and baselines in different label proportions are shown in Table \ref{tab: Effects of label magnitude}. Except for the comparable classification accuracy on WikiCS at a 0.5$\%$ label ratio, iGCL achieves the optimal classification accuracies for any label ratio. We find that the classification accuracies of DGI are significantly weaker than those of supervised GNNs such as GCN and GAT in the absence of labels. As the labels are gradually enriched, the classification accuracy gap between DGI and GNNs is gradually reduced. When the label ratio is 20$\%$, the classification accuracy of DGI is basically comparable to that of GNNs. This indicates that DGI can achieve excellent performance only when the labels are sufficient. Compared with DGI, iGCL has better generalizability to adapt different proportions of labels. The classification accuracies of iGCL always exceed those of GNNs regardless of label sufficiency or scarcity. The experimental results show that iGCL has excellent generalization.

\begin{table}
    \centering
    \caption{Summary of statistical results in terms of mean accuracies and standard deviation under different label ratios. \textbf{Bold} numbers indicate best results.}
    \label{tab: Effects of label magnitude}
    \setlength{\tabcolsep}{0.5mm}{
    \begin{tabular}{ccccccc}
        \toprule
         & 0.5$\%$ & 1$\%$ & 2$\%$ & 5$\%$ & 10$\%$ & 20$\%$\\ 
        \midrule
        GCN.C & 77.85Â±2.2 & 81.87Â±2.0 & 85.08Â±0.8 & 87.03Â±0.6 & 87.70Â±0.4 & 88.11Â±0.3 \\
        GAT.C & 76.41Â±3.0 & 79.88Â±2.5 & 83.22Â±1.5 & 85.37Â±1.1 & 86.13Â±0.8 & 86.38Â±0.9 \\
        DGI.C & 72.33Â±1.9 & 74.89Â±1.3 & 78.57Â±1.0 & 84.50Â±0.5 & 87.48Â±0.4 & 88.43Â±0.3 \\
        iGCL.C & \textbf{78.20Â±1.7} & \textbf{83.01Â±1.5} & \textbf{86.20Â±0.7} & \textbf{88.51Â±0.6} & \textbf{89.79Â±0.3} & \textbf{90.52Â±0.2} \\
        \hline
         GCN.W & 62.22Â±3.7 & 68.06Â±2.9 & 72.08Â±1.6 & 75.60Â±0.9 & 76.65Â±0.6 & 77.63Â±0.3 \\
         GAT.W & \textbf{62.33Â±3.8} & 67.57Â±2.2 & 71.20Â±1.6 & 74.35Â±1.1 & 75.90Â±0.6 & 76.61Â±0.5 \\
         DGI.W & 54.65Â±3.2 & 61.04Â±2.4 & 66.59Â±1.7 & 71.15Â±0.9 & 75.14Â±0.4 & 77.32Â±0.4 \\
         iGCL.W & 61.85Â±3.2 & \textbf{70.04Â±1.8} & \textbf{74.48Â±1.3} & \textbf{77.32Â±0.6} & \textbf{79.66Â±0.4} & \textbf{81.18Â±0.4} \\
        \bottomrule
    \end{tabular}
    }
\end{table}

\subsection{Robustness analysis}

We evaluate the robustness of iGCL by the classification accuracies on the attacked datasets. We choose Nettack \cite{zugner2018adversarial} as the graph attack method. Nettack can attack structure or node features. Following the experimental protocol, we choose the first division of CS and WikiCS as the division for the subsection. We utilize only the training and test sets and discard the validation sets. Since Nettack cannot compute in parallel and requires considerable time, we attack one-tenth of the nodes in the test sets. We train models in the poison way, i.e., we train models with the attacked datasets from scratch. The performance of the iGCL and baselines under different types of attacks is shown in Table \ref{tab: Robustness analysis}. The classification accuracies of all models degrade after the attacks, which indicates that Nettack can indeed degrade the performance. Under any type of attack, the classification accuracies of iGCL stably exceed those of the baselines. In particular, on the WikiCS dataset, the classification accuracy of iGCL is significantly higher than those of the baselines. The experimental results show that iGCL has excellent robustness and has the ability to resist graph attack algorithms.

\begin{table}[h]
    \centering
    \caption{Summary of statistical results in terms of mean accuracies and standard deviation under different label ratios. \textbf{Bold} numbers indicate best results.}
    \label{tab: Robustness analysis}
    \setlength{\tabcolsep}{2mm}{
    \begin{tabular}{ccccc}
        \toprule
         & None & Struc. & Feats. & Struc. $\&$ Feats. \\ 
        \midrule
        GCN.C & 92.55Â±0.1 & 83.61Â±0.1 & 91.98Â±0.0 & 83.58Â±0.1\\
        GAT.C & 92.40Â±0.1 & 82.13Â±0.1 & 90.95Â±0.2 & 82.16Â±0.1\\
        DGI.C & 92.38Â±0.2 & 84.22Â±0.1 & 92.19Â±0.1 & 84.21Â±0.1\\
        iGCL.C & \textbf{93.01Â±0.1} & \textbf{84.48Â±0.0} & \textbf{92.27Â±0.0} & \textbf{84.50Â±0.0}\\
        \hline
         GCN.W & 76.83Â±0.2 & 36.88Â±0.3 & 39.62Â±0.4 & 33.99Â±0.5 \\
         GAT.W & 76.55Â±0.3 & 35.82Â±1.2 & 32.28Â±1.9 & 30.76Â±2.1\\
         DGI.W & 74.22Â±0.4 & 34.44Â±0.0 & 35.97Â±0.1 & 33.14Â±0.1\\
         iGCL.W & \textbf{78.09Â±0.6} & \textbf{41.91Â±0.1} & \textbf{42.15Â±0.1} & \textbf{40.86Â±0.5}\\
        \bottomrule
    \end{tabular}
    }
\end{table}

\subsection{Visualization of representations}

\begin{figure}[h]
  \begin{center}
  \includegraphics[width=3in]{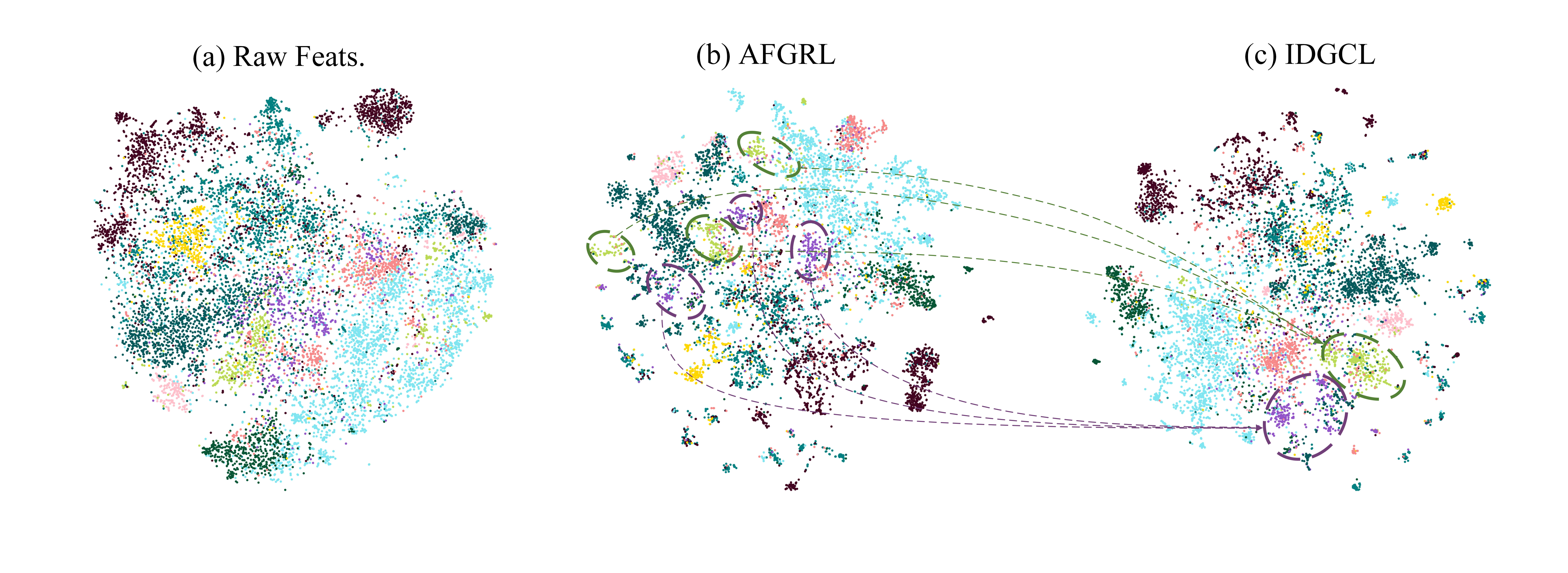}
  \end{center}
  \caption{Visualization of node representations of WikiCS by $t$-SNE.}
  \label{fig:t-sne}
\end{figure}

We visualize the raw features and node representations learned by AFGRL and iGCL by $t$-SNE, thus providing an intuitive understanding of the learned representations. The visualization of node representations is shown in Fig. \ref{fig:t-sne}. Each point represents a node, and the node color represents the class of the node. We select the silhouette coefficient to evaluate the overall clustering results. The larger the silhouette coefficient, the better the results. The silhouette coefficients of the raw features, AFGRL, and iGCL are -0.0944, -0.0516, and -0.0322, respectively. The silhouette coefficients of AFGRL and iGCL are larger than that of raw feats, which indicates that both AFGRL and iGCL can effectively improve the ability to cluster the same classes of nodes and separate different classes of nodes. Furthermore, iGCL has a stronger ability than AFGRL. For example, iGCL aggregates the light green nodes and purple nodes into the same regions, while AFGCL aggregates these nodes into different regions, as shown in the oval region in Fig. \ref{fig:t-sne}. This indicates that the iGCL can effectively aggregate similar nodes and discriminate dissimilar nodes.

\subsection{Hyperparameter sensitivity analysis}

We analyze the impact of four key hyperparameters in detail: balance coefficient $\lambda$, dimension of representations $D$, number of graph neural layers $L$, and number of positive samples $K$.

\begin{figure}[h]
  \begin{center}
  \includegraphics[width=3in]{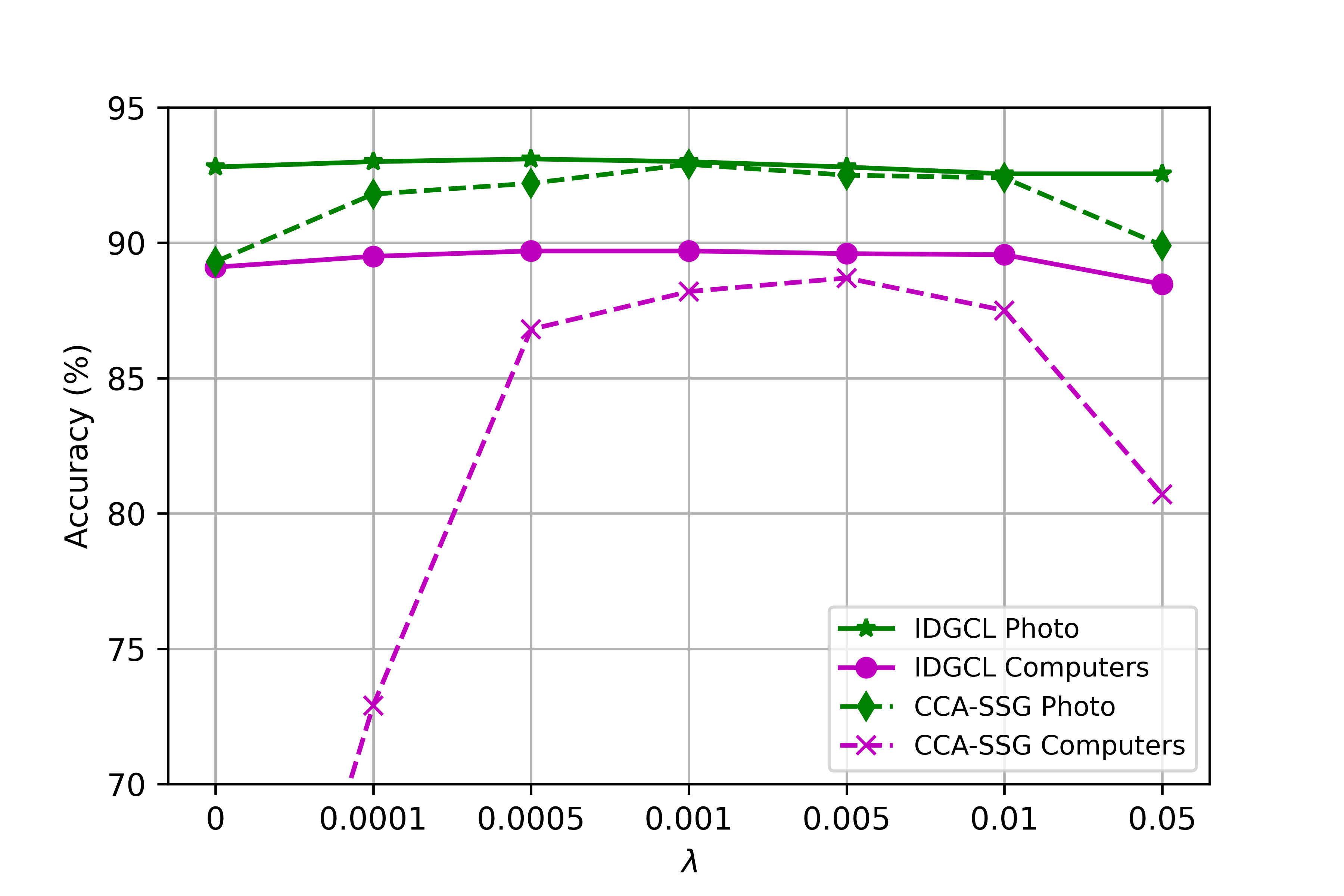}
  \end{center}
  \caption{Effects of balance coefficient $\lambda$ for iGCL and CCA-SSG.}
  \label{fig:lambda}
\end{figure}

The effect of the balance coefficient $\lambda$. We theoretically prove that the ID loss and the loss of CCA-SSG are equivalent. For different $\lambda$, the classification accuracies of iGCL and CCA-SSG on Photo and Computers are shown in Fig. \ref{fig:lambda}. When $\lambda = 0$, i.e., the discrimination term is removed, the classification accuracies of both iGCL and CCA-SSG decrease compared to the optimal classification accuracies. This indicates that the discrimination term can effectively improve the quality of the representations. Note that the performance of iGCL is only weakly degraded, while the performance of CCA-SSG is significantly degraded. In particular, the classification accuracy of CCA-SSG for Computers is only 53.7$\%$, which implies a trivial solution. Regardless of whether $\lambda$ is small or large, the magnitude of the performance variation of iGCL is always tiny, while CCA-SSG requires a suitable $\lambda$ to ensure the performance. This indicates that iGCL is much less sensitive to $\lambda$ than CCA-SSG.

\begin{figure}[h]
  \begin{center}
  \includegraphics[width=3in]{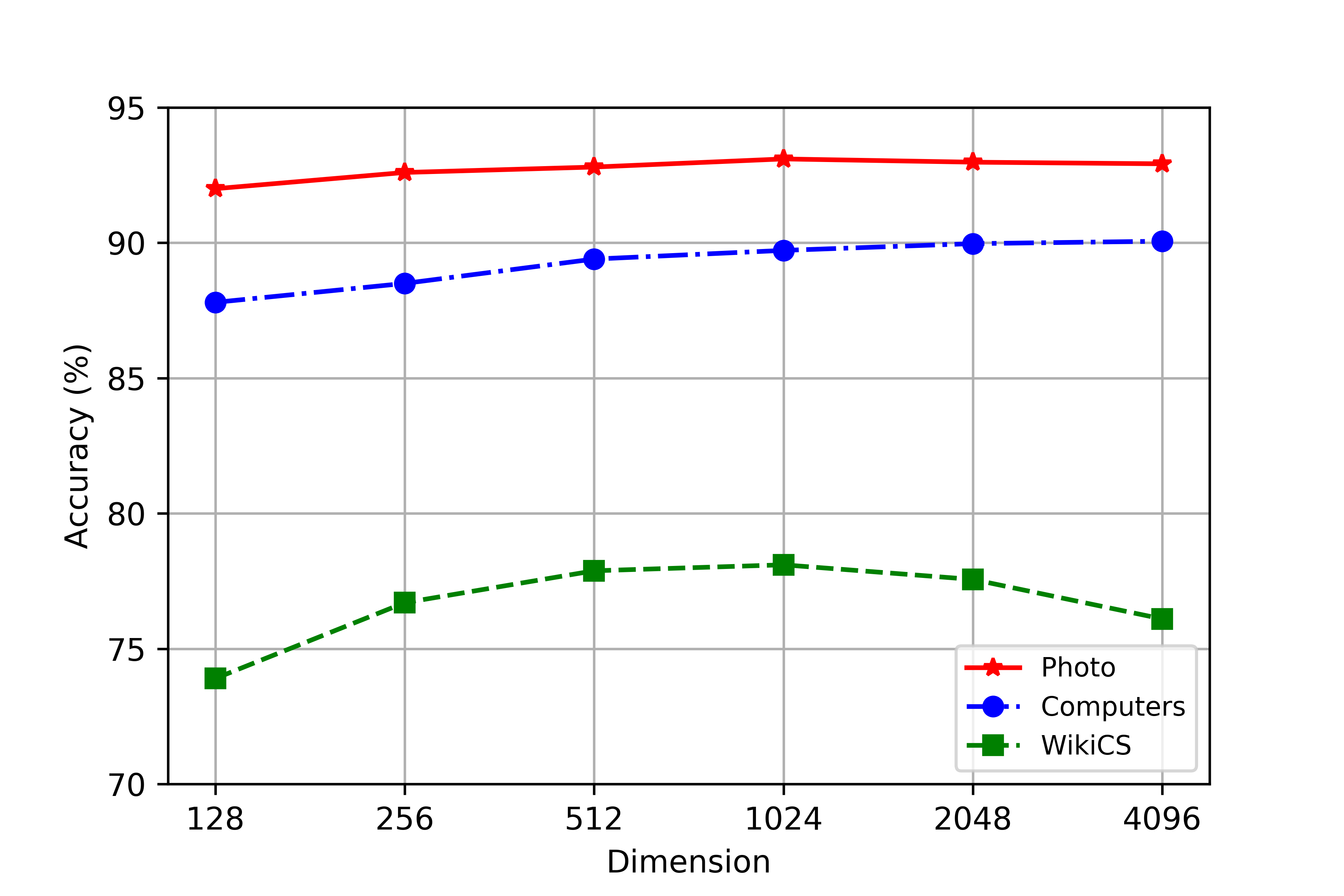}
  \end{center}
  \caption{Effects of the dimension of representations $D$ for iGCL.}
  \label{fig: dimension of representation}
\end{figure}

We analyze the effects of $D$ on the Photo and Computers, as shown in Fig. \ref{fig: dimension of representation}. When $D=1024$, iGCL achieves satisfactory classification accuracies on most datasets. This is different from unsupervised graph representation learning methods such as node2vec, which usually set $D$ to 128. Moreover, $D=1024$ is also higher than those of GCL methods such as DGI and CCA-SSG, which are usually 512. Note that the simple logistic regression classifier can correctly classify the representations even if $D$ is as high as 1024. Moreover, the performance of iGCL on Photo and Computers does not show significant degradation when $D$ is further increased. In particular, for Computers, iGCL achieves the best classification accuracy when $D$ is 4096. This indicates that the high-dimensional representations generated by the iGCL can enhance its discriminative power and that the curse of dimensionality does not occur. Since increasing $D$ increases the computational memory, we suggest setting the default $D$ to 1024.

\begin{figure}[h]
  \begin{center}
  \includegraphics[width=3in]{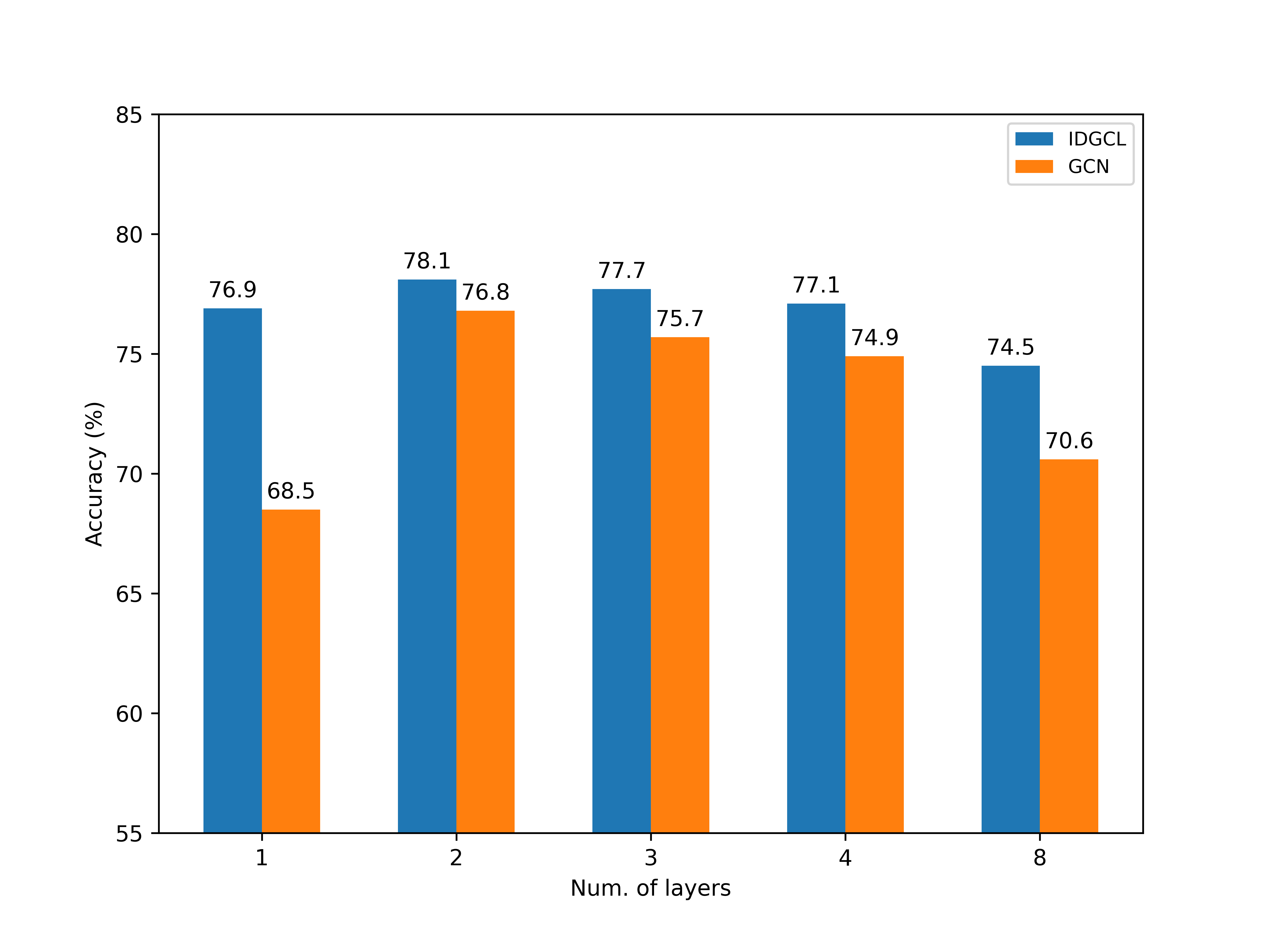}
  \end{center}
  \caption{The effects of the number of graph neural layers $L$ for iGCL.}
  \label{fig: number of graph neural layers}
\end{figure}

We explore the effect of $L$ on the performance of iGCL. The classification accuracies of iGCL and GCN in WikiCS with different $L$ are shown in Fig. \ref{fig: number of graph neural layers}. We find that the classification accuracy of iGCL is highest when $L$ is 2. This indicates that iGCL prefers shallow networks. Then, the classification accuracies of iGCL decrease with an increasing $L$. The experimental results show that the oversmoothing of GNNs also exists in iGCL. We notice that the classification accuracies of GCN decrease significantly when $L$ is 3 or 4. This indicates that the oversmoothing of GCN is very serious. For iGCL, the classification accuracy shows a significant decline only when $L$ reaches 8, and the decline is also lower than that of GCN. This indicates that iGCL can alleviate the oversmoothing problem.

\begin{figure}[h]
  \begin{center}
  \includegraphics[width=3in]{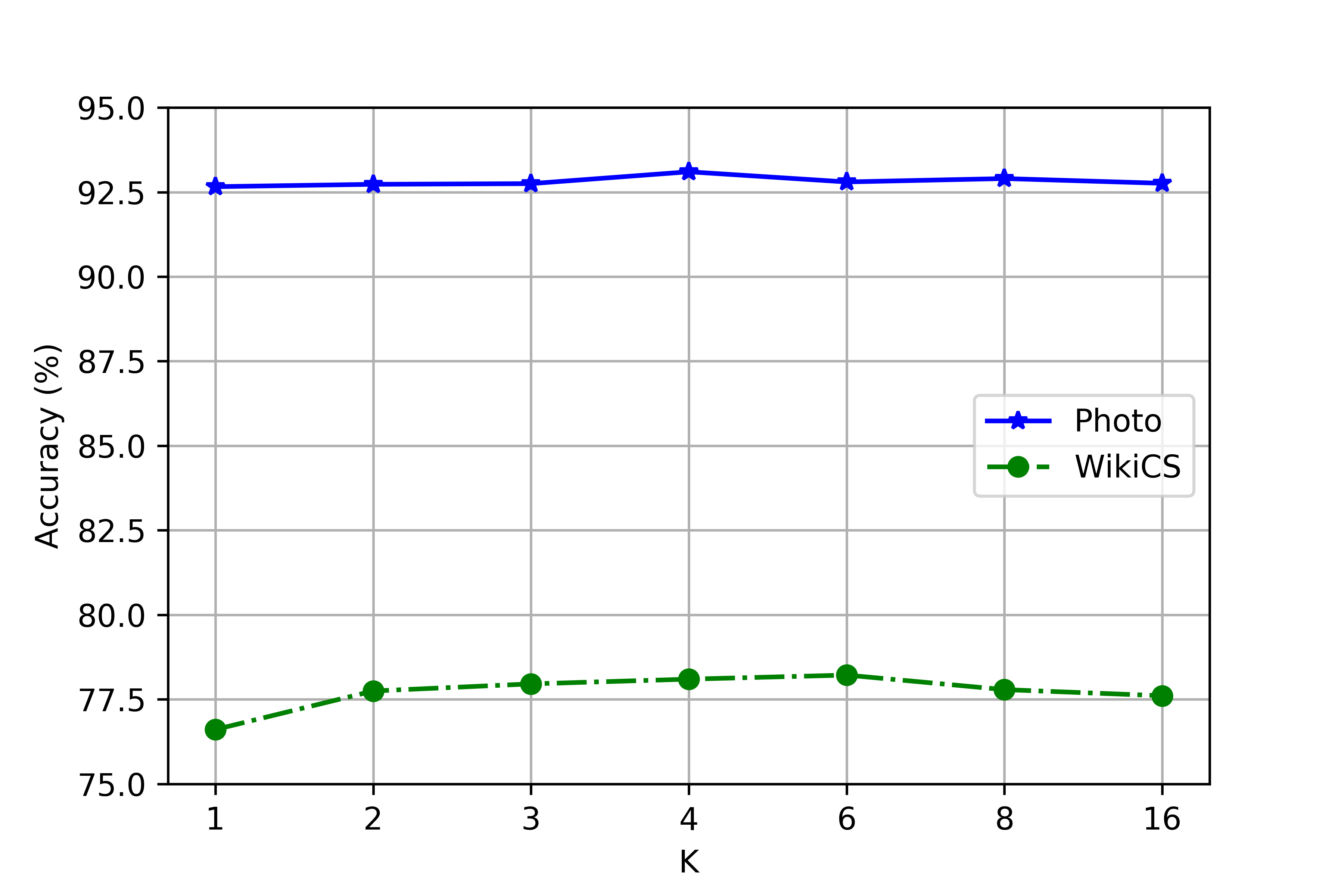}
  \end{center}
  \caption{Effects of the number of positive samples $K$ for iGCL.}
  \label{fig: number of positive samples}
\end{figure}

We analyze the effect of $K$ on the performance of iGCL. $K=1$ means that only $H_i^{\xi}$ of the target node $i$ is chosen as the positive sample. We measure the classification accuracies of iGCL on Photo and WikiCS when $K$ takes different values, as shown in Fig. \ref{fig: number of positive samples}. Compared with $K=1$, there are always gains when using neighboring nodes as positive samples. In particular, the classification accuracy at $K=6$ improves by 1.5$\%$ over that at $K=1$ on WikiCS. The experimental results indicate that the positive sample construction strategy is effective and makes full use of the proximity. We also notice a small decrease in the classification accuracies when $K$ is too large, such as $K=16$. The reason is that some nodes with different classes dilute the useful and invariant information. AFGRL requires $k$-nn and $k$-means clustering algorithms for all nodes to filter positive samples, which requires tremendous computational resources and time. Because the $k$-nn and $k$-means algorithms require all representations, AFGRL is unsuitable for large-scale datasets. In contrast, iGCL can naturally be applied to large-scale datasets. DGCL only requires the neighboring nodes of target nodes and can be trained in a minibatch manner with the help of neighbor sampling techniques \cite{hamilton2017inductive}.

\section{Conclusion}
In this paper, we proposed an efficient and effective augmentation-free GCL method called invariant-discriminative graph contrastive learning (iGCL). To avoid empirical data augmentation, iGCL generates an augmented view of the entire graph through Siamese networks. Then, the positive sample construction strategy can effectively select nodes with similar semantic information to the target node from neighboring nodes of the augmented view as positive samples. This enriches the diversity of positive samples and improves the performance of iGCL. Different from instance-level mutual information estimation, the ID loss is at the feature level. ID loss selects the MSE as the invariance term, which forces the positive samples to be as similar as possible to the target sample in the representation space. ID loss uses the orthonormal constraint as the discrimination term to force the representations in different dimensions to be independent of each other and to guarantee the discriminability of representations.  

Extensive experimental results showed that iGCL outperforms all baselines on 5 node classification benchmark datasets. Using the representations generated by iGCL as input, the simple logistic regression can outperform the complex GNNs. iGCL also shows superior performance for different label ratios and is capable of resisting graph attacks, which indicates that iGCL has excellent generalization and robustness. A visualization experiment illustrated the ability of iGCL to aggregate nodes of the same kind and separate nodes of different classes without relying on manual labeling. Parameter sensitivity experiments illustrated that the performance of the iGCL is insensitive to the values of coefficient   and number of positive samples. Although iGCL is suitable for shallow GCNs, it shows better resistance to oversmoothing than GCNs. Intriguingly, iGCL benefits from very high-dimensional representations, similar to the findings of \cite{zbontar2021barlow}. Since iGCL is based on BYOL, it employs tricks such as an asymmetric structure and additional projector to avoid trivial solutions. Improving the network architecture of iGCL deserves further exploration.

\section{ACKNOWLEDGMENT}
The authors would like to thank anonymous reviewers for their valuable suggestions to make this article better. They would also like to thank the High Performance Computing Platform of Central South University for providing high performance computing (HPC) resources.


\bibliographystyle{IEEEtran}
\bibliography{refs}

\begin{thebibliography}{10}
\providecommand{\url}[1]{#1}
\csname url@samestyle\endcsname
\providecommand{\newblock}{\relax}
\providecommand{\bibinfo}[2]{#2}
\providecommand{\BIBentrySTDinterwordspacing}{\spaceskip=0pt\relax}
\providecommand{\BIBentryALTinterwordstretchfactor}{4}
\providecommand{\BIBentryALTinterwordspacing}{\spaceskip=\fontdimen2\font plus
\BIBentryALTinterwordstretchfactor\fontdimen3\font minus
  \fontdimen4\font\relax}
\providecommand{\BIBforeignlanguage}[2]{{%
\expandafter\ifx\csname l@#1\endcsname\relax
\typeout{** WARNING: IEEEtran.bst: No hyphenation pattern has been}%
\typeout{** loaded for the language `#1'. Using the pattern for}%
\typeout{** the default language instead.}%
\else
\language=\csname l@#1\endcsname
\fi
#2}}
\providecommand{\BIBdecl}{\relax}
\BIBdecl

\bibitem{hjelm2018learning}
R.~D. Hjelm, A.~Fedorov, S.~Lavoie-Marchildon, K.~Grewal, P.~Bachman,
  A.~Trischler, and Y.~Bengio, ``Learning deep representations by mutual
  information estimation and maximization,'' in \emph{International Conference
  on Learning Representations}, 2018.

\bibitem{you2020graph}
Y.~You, T.~Chen, Y.~Sui, T.~Chen, Z.~Wang, and Y.~Shen, ``Graph contrastive
  learning with augmentations,'' \emph{Advances in Neural Information
  Processing Systems}, vol.~33, pp. 5812--5823, 2020.

\bibitem{chen2020simple}
T.~Chen, S.~Kornblith, M.~Norouzi, and G.~Hinton, ``A simple framework for
  contrastive learning of visual representations,'' in \emph{International
  conference on machine learning}.\hskip 1em plus 0.5em minus 0.4em\relax PMLR,
  2020, pp. 1597--1607.

\bibitem{liu2022graph}
Y.~Liu, M.~Jin, S.~Pan, C.~Zhou, Y.~Zheng, F.~Xia, and P.~Yu, ``Graph
  self-supervised learning: A survey,'' \emph{IEEE Transactions on Knowledge
  and Data Engineering}, 2022.

\bibitem{zhu2020deep}
Y.~Zhu, Y.~Xu, F.~Yu, Q.~Liu, S.~Wu, and L.~Wang, ``Deep graph contrastive
  representation learning,'' \emph{arXiv preprint arXiv:2006.04131}, 2020.

\bibitem{zhang2021canonical}
H.~Zhang, Q.~Wu, J.~Yan, D.~Wipf, and P.~S. Yu, ``From canonical correlation
  analysis to self-supervised graph neural networks,'' \emph{Advances in Neural
  Information Processing Systems}, vol.~34, pp. 76--89, 2021.

\bibitem{hassani2020contrastive}
K.~Hassani and A.~H. Khasahmadi, ``Contrastive multi-view representation
  learning on graphs,'' in \emph{International Conference on Machine
  Learning}.\hskip 1em plus 0.5em minus 0.4em\relax PMLR, 2020, pp. 4116--4126.

\bibitem{page1999pagerank}
L.~Page, S.~Brin, R.~Motwani, and T.~Winograd, ``The pagerank citation ranking:
  Bringing order to the web.'' Stanford InfoLab, Tech. Rep., 1999.

\bibitem{kondor2002diffusion}
R.~I. Kondor and J.~Lafferty, ``Diffusion kernels on graphs and other discrete
  structures,'' in \emph{Proceedings of the 19th international conference on
  machine learning}, vol. 2002, 2002, pp. 315--322.

\bibitem{jiao2020sub}
Y.~Jiao, Y.~Xiong, J.~Zhang, Y.~Zhang, T.~Zhang, and Y.~Zhu, ``Sub-graph
  contrast for scalable self-supervised graph representation learning,'' in
  \emph{2020 IEEE international conference on data mining (ICDM)}.\hskip 1em
  plus 0.5em minus 0.4em\relax IEEE, 2020, pp. 222--231.

\bibitem{velickovic2019deep}
P.~Velickovic, W.~Fedus, W.~L. Hamilton, P.~Li{\`o}, Y.~Bengio, and R.~D.
  Hjelm, ``Deep graph infomax.'' \emph{ICLR (Poster)}, vol.~2, no.~3, p.~4,
  2019.

\bibitem{wang2020understanding}
T.~Wang and P.~Isola, ``Understanding contrastive representation learning
  through alignment and uniformity on the hypersphere,'' in \emph{International
  Conference on Machine Learning}.\hskip 1em plus 0.5em minus 0.4em\relax PMLR,
  2020, pp. 9929--9939.

\bibitem{bromley1993signature}
J.~Bromley, I.~Guyon, Y.~LeCun, E.~S{\"a}ckinger, and R.~Shah, ``Signature
  verification using a" siamese" time delay neural network,'' \emph{Advances in
  neural information processing systems}, vol.~6, 1993.

\bibitem{kefato2021jointly}
Z.~T. Kefato, S.~Girdzijauskas, and H.~St{\"a}rk, ``Jointly learnable data
  augmentations for self-supervised gnns,'' \emph{arXiv preprint
  arXiv:2108.10420}, 2021.

\bibitem{mcpherson2001birds}
M.~McPherson, L.~Smith-Lovin, and J.~M. Cook, ``Birds of a feather: Homophily
  in social networks,'' \emph{Annual review of sociology}, pp. 415--444, 2001.

\bibitem{grover2016node2vec}
A.~Grover and J.~Leskovec, ``node2vec: Scalable feature learning for
  networks,'' in \emph{Proceedings of the 22nd ACM SIGKDD international
  conference on Knowledge discovery and data mining}, 2016, pp. 855--864.

\bibitem{he2020momentum}
K.~He, H.~Fan, Y.~Wu, S.~Xie, and R.~Girshick, ``Momentum contrast for
  unsupervised visual representation learning,'' in \emph{Proceedings of the
  IEEE/CVF conference on computer vision and pattern recognition}, 2020, pp.
  9729--9738.

\bibitem{rottmann2020detection}
M.~Rottmann, K.~Maag, R.~Chan, F.~H{\"u}ger, P.~Schlicht, and H.~Gottschalk,
  ``Detection of false positive and false negative samples in semantic
  segmentation,'' in \emph{2020 Design, Automation \& Test in Europe Conference
  \& Exhibition (DATE)}.\hskip 1em plus 0.5em minus 0.4em\relax IEEE, 2020, pp.
  1351--1356.

\bibitem{grill2020bootstrap}
J.-B. Grill, F.~Strub, F.~Altch{\'e}, C.~Tallec, P.~Richemond, E.~Buchatskaya,
  C.~Doersch, B.~Avila~Pires, Z.~Guo, M.~Gheshlaghi~Azar \emph{et~al.},
  ``Bootstrap your own latent-a new approach to self-supervised learning,''
  \emph{Advances in neural information processing systems}, vol.~33, pp.
  21\,271--21\,284, 2020.

\bibitem{ren2019heterogeneous}
Y.~Ren, B.~Liu, C.~Huang, P.~Dai, L.~Bo, and J.~Zhang, ``Heterogeneous deep
  graph infomax,'' \emph{arXiv preprint arXiv:1911.08538}, 2019.

\bibitem{li2021leveraging}
X.~Li, D.~Ding, B.~Kao, Y.~Sun, and N.~Mamoulis, ``Leveraging meta-path
  contexts for classification in heterogeneous information networks,'' in
  \emph{2021 IEEE 37th International Conference on Data Engineering
  (ICDE)}.\hskip 1em plus 0.5em minus 0.4em\relax IEEE, 2021, pp. 912--923.

\bibitem{zhu2021graph}
Y.~Zhu, Y.~Xu, F.~Yu, Q.~Liu, S.~Wu, and L.~Wang, ``Graph contrastive learning
  with adaptive augmentation,'' in \emph{Proceedings of the Web Conference
  2021}, 2021, pp. 2069--2080.

\bibitem{thakoor2021bootstrapped}
S.~Thakoor, C.~Tallec, M.~G. Azar, R.~Munos, P.~Veli{\v{c}}kovi{\'c}, and
  M.~Valko, ``Bootstrapped representation learning on graphs,'' in \emph{ICLR
  2021 Workshop on Geometrical and Topological Representation Learning}, 2021.

\bibitem{lee2022augmentation}
N.~Lee, J.~Lee, and C.~Park, ``Augmentation-free self-supervised learning on
  graphs,'' in \emph{Proceedings of the AAAI Conference on Artificial
  Intelligence}, vol.~36, no.~7, 2022, pp. 7372--7380.

\bibitem{belkin2003laplacian}
M.~Belkin and P.~Niyogi, ``Laplacian eigenmaps for dimensionality reduction and
  data representation,'' \emph{Neural computation}, vol.~15, no.~6, pp.
  1373--1396, 2003.

\bibitem{barlow1961possible}
H.~B. Barlow \emph{et~al.}, ``Possible principles underlying the transformation
  of sensory messages,'' \emph{Sensory communication}, vol.~1, no.~01, 1961.

\bibitem{zbontar2021barlow}
J.~Zbontar, L.~Jing, I.~Misra, Y.~LeCun, and S.~Deny, ``Barlow twins:
  Self-supervised learning via redundancy reduction,'' in \emph{International
  Conference on Machine Learning}.\hskip 1em plus 0.5em minus 0.4em\relax PMLR,
  2021, pp. 12\,310--12\,320.

\bibitem{andrew2013deep}
G.~Andrew, R.~Arora, J.~Bilmes, and K.~Livescu, ``Deep canonical correlation
  analysis,'' in \emph{International conference on machine learning}.\hskip 1em
  plus 0.5em minus 0.4em\relax PMLR, 2013, pp. 1247--1255.

\bibitem{gong2014multi}
Y.~Gong, Q.~Ke, M.~Isard, and S.~Lazebnik, ``A multi-view embedding space for
  modeling internet images, tags, and their semantics,'' \emph{International
  journal of computer vision}, vol. 106, no.~2, pp. 210--233, 2014.

\bibitem{chang2018scalable}
X.~Chang, T.~Xiang, and T.~M. Hospedales, ``Scalable and effective deep cca via
  soft decorrelation,'' in \emph{Proceedings of the IEEE Conference on Computer
  Vision and Pattern Recognition}, 2018, pp. 1488--1497.

\bibitem{amjad2019learning}
R.~A. Amjad and B.~C. Geiger, ``Learning representations for neural
  network-based classification using the information bottleneck principle,''
  \emph{IEEE transactions on pattern analysis and machine intelligence},
  vol.~42, no.~9, pp. 2225--2239, 2019.

\bibitem{federici2020learning}
M.~Federici, A.~Dutta, P.~Forr{\'e}, N.~Kushman, and Z.~Akata, ``Learning
  robust representations via multi-view information bottleneck,'' \emph{arXiv
  preprint arXiv:2002.07017}, 2020.

\bibitem{sinha2015overview}
A.~Sinha, Z.~Shen, Y.~Song, H.~Ma, D.~Eide, B.-J. Hsu, and K.~Wang, ``An
  overview of microsoft academic service (mas) and applications,'' in
  \emph{Proceedings of the 24th international conference on world wide web},
  2015, pp. 243--246.

\bibitem{mcauley2015image}
J.~McAuley, C.~Targett, Q.~Shi, and A.~Van Den~Hengel, ``Image-based
  recommendations on styles and substitutes,'' in \emph{Proceedings of the 38th
  international ACM SIGIR conference on research and development in information
  retrieval}, 2015, pp. 43--52.

\bibitem{mernyei2020wiki}
P.~Mernyei and C.~Cangea, ``Wiki-cs: A wikipedia-based benchmark for graph
  neural networks,'' \emph{arXiv preprint arXiv:2007.02901}, 2020.

\bibitem{kipf2016semi}
T.~N. Kipf and M.~Welling, ``Semi-supervised classification with graph
  convolutional networks,'' \emph{arXiv preprint arXiv:1609.02907}, 2016.

\bibitem{velivckovic2017graph}
P.~Veli{\v{c}}kovi{\'c}, G.~Cucurull, A.~Casanova, A.~Romero, P.~Lio, and
  Y.~Bengio, ``Graph attention networks,'' \emph{arXiv preprint
  arXiv:1710.10903}, 2017.

\bibitem{hamilton2017inductive}
W.~Hamilton, Z.~Ying, and J.~Leskovec, ``Inductive representation learning on
  large graphs,'' \emph{Advances in neural information processing systems},
  vol.~30, 2017.

\bibitem{zugner2018adversarial}
D.~Z{\"u}gner, A.~Akbarnejad, and S.~G{\"u}nnemann, ``Adversarial attacks on
  neural networks for graph data,'' in \emph{Proceedings of the 24th ACM SIGKDD
  international conference on knowledge discovery \& data mining}, 2018, pp.
  2847--2856.

\end{thebibliography}

\end{document}